\pdfoutput=1

\documentclass[11pt]{article}

\usepackage[final]{acl}

\providecommand{\linenumbers}{}

\usepackage{times}
\usepackage{latexsym}
\usepackage{amsmath}
\usepackage{amssymb}
\usepackage{graphicx}
\usepackage{booktabs}
\usepackage{multirow}
\usepackage{float}
\usepackage{microtype}
\usepackage{xcolor}
\usepackage{listings}
\usepackage{algorithm}
\usepackage{algorithmic}
\usepackage{tikz}
\usetikzlibrary{shapes.geometric, arrows.meta, positioning, fit, backgrounds}

\title{ChunkRank: Model-Aware Text Chunking and Abstention-Aware\\
  Answer Selection for LLM Pipelines}

\author{
  Amit Nautiyal$^{\href{https://orcid.org/0009-0003-5298-1453}{\textsuperscript{iD}}}$ \\
  \footnotesize\texttt{research.amit.n@gmail.com} \\\And
  Ayush Bhatt$^{\href{https://orcid.org/0009-0001-1060-9253}{\textsuperscript{iD}}}$ \\
  \footnotesize\texttt{ayushbhatt1224@gmail.com} \\\And
  Gaurav Nautiyal$^{\href{https://orcid.org/0009-0003-6132-1692}{\textsuperscript{iD}}}$ \\
  \footnotesize\texttt{26021892@geu.ac.in}
}

\begin{document}

\maketitle

\begin{abstract}
We present \textbf{ChunkRank}, an open-source Python library that derives chunk
boundaries from a target model's tokenizer and context window, and selects an
answer among candidates produced independently per chunk. It ships a validated
registry of 90 models across 15 providers and six answer-selection methods, and
needs only three core dependencies. For chunking, ChunkRank avoids context-window
overflow automatically from the model name, whereas character-based splitters
overflow or waste the budget, and a fidelity study across 11 languages shows why
token-exact budgets matter beyond English. For answer selection we report a
negative result: on NaturalQuestions, TriviaQA and HotpotQA, with extractive and
generative readers, no content-based ranker reliably beats taking the first
non-empty answer. The reason is reader abstention on chunks that lack the answer,
not answer position. A long-context baseline shows that chunking matches
single-call reading on single-hop questions, so ChunkRank targets small-window and
beyond-window settings. Code, registry and evaluation harness are released.
\end{abstract}

\section{Introduction}
\label{sec:intro}

All deployed large language models accept inputs up to some fixed token
budget. Although this limit has grown considerably in recent years (from 2{,}048
tokens in GPT-3 \cite{brown2020gpt3} to 1M tokens in Gemini 1.5
\cite{gemini15report} and 10M tokens in Llama 4 Scout \cite{meta2025llama4}),
documents such as legal contracts, research corpora, codebases, and
multi-chapter reports routinely exceed even the largest current windows.
Long documents must therefore be split before they can be processed, and
every practitioner working on such documents needs a reliable, reusable
component for doing so.

The NLP community has a long tradition of releasing general-purpose toolkits
that lower the barrier to reproducible engineering (Section~\ref{sec:related}),
but chunking for LLM pipelines has not received the same treatment. Existing
splitters \cite{chase2022langchain,liu2022llamaindex,haystack2019deepset} bundle
character- or token-based logic as one utility inside a larger retrieval
framework, requiring a user-supplied chunk size and never consulting the target
model's tokenizer. Because tokenization varies across model families, a
512-character window is anywhere from 100 to 400 tokens depending on the encoder
\cite{sennrich2016bpe,kudo2018sentencepiece}, so a size tuned for one model
silently truncates or wastes context on another, a mismatch practitioners
discover through degraded output rather than explicit errors, and, absent a
standalone model-aware resource, re-solve independently and inconsistently.

Chunking introduces a second, less-discussed challenge: \emph{answer
multiplicity}. Queried independently against each chunk, the LLM may produce a
different answer per segment; most applications simply return the first chunk's,
an arbitrary and empirically suboptimal choice. RAG systems
\cite{lewis2020rag,izacard2021fid,guu2020realm} address a related problem by
ranking retrieved passages before generation, but within a monolithic pipeline
that conflates retrieval, indexing, chunking, and generation, and not as a
standalone resource a non-RAG pipeline can adopt.

We introduce \textbf{ChunkRank} to close this resource gap: an installable
package (not a code snippet embedded in a larger framework) with a
versioned model registry, a documented and stable API across three levels
of abstraction, a public test suite exercised on four Python versions,
and a permissive license. This paper documents the resource's design, the
engineering decisions that make it adoptable in constrained environments,
and an empirical study showing that answer selection over chunks is dominated by
reader abstention, so a trivial first-non-empty rule is hard to beat, with real
oracle headroom that content-based ranking does not capture.

This paper makes the following contributions:

\begin{enumerate}
  \item We release ChunkRank, an open-source, versioned resource that
        packages model-aware chunk boundary computation and post-chunk answer
        ranking as reusable, independently documented components
        (Sections~\ref{sec:design}--\ref{sec:impl}).

  \item We release a community-extensible \emph{model registry} of 90
        pre-configured models across 15 providers, with a documented
        schema and a runtime registration API so that new models can be
        added without a library upgrade (Section~\ref{sec:registry}).

  \item We document the resource's reproducibility guarantees: license,
        dependency footprint, supported Python versions, and public test
        coverage, to support long-term reuse by the community
        (Section~\ref{sec:resource}).

  \item We validate the resource empirically and report a negative result about
        answer selection: across NaturalQuestions, TriviaQA, and HotpotQA, with
        both an extractive and a generative reader and paired bootstrap tests, no
        content-based ranker reliably beats taking the first non-empty answer,
        because selection over chunks is dominated by the reader's ability to
        abstain (Section~\ref{sec:experiments}). We adopt first-non-empty as the
        library default, add a long-context baseline that delimits when chunking
        is needed at all, and release the full harness, configuration, and result
        tables.
\end{enumerate}

\paragraph{Downstream impact.} The setting ChunkRank targets is increasingly
common: tool-using agents and multi-step reasoning chains that feed long,
externally-retrieved documents (web pages, code, PDFs, tool outputs) into a model
with a fixed budget, often one whose tokenizer is not the developer's own. There,
a silent overflow surfaces as a truncated tool result or degraded answer several
steps downstream, where it is hard to attribute. By deriving safe boundaries from
the named model and turning implicit first-chunk selection into an explicit,
swappable answer-selection step, ChunkRank makes both failure modes explicit at the point
of chunking, and as model-agnostic middleware drops into an agent loop, an offline
document-processing workflow, or a RAG system without a pipeline rewrite.

ChunkRank is available under the Apache 2.0 license on PyPI at
\url{https://pypi.org/project/chunkrank/} and on GitHub at
\url{https://github.com/AmitoVrito/chunkrank}, where the source, model
registry JSON, and test suite are publicly browsable.

\section{Related Work}
\label{sec:related}

\subsection{NLP Toolkits and Resource Papers}

The NLP community has a long history of resource and tool papers
that package research-grade functionality for reuse: NLTK \cite{bird2009nltk},
spaCy \cite{honnibal2020spacy}, Stanza \cite{qi2020stanza}, and Flair
\cite{akbik2019flair} package tokenization, tagging, and parsing behind stable,
versioned APIs, and Hugging Face Transformers \cite{wolf2020transformers}
generalized the pattern to pretrained-model access. ChunkRank follows this
tradition but fills a gap none of these do: adapting segmentation to the
tokenizer and context budget of a \emph{named, deployed LLM}, and resolving the
resulting answer-selection problem, as a standalone, dependency-light component.

\subsection{Text Chunking in LLM Pipelines}

Token-budget management for LLMs is discussed informally in practitioner
documentation but has received little formal treatment as a resource
problem. LangChain's recursive character splitter \cite{chase2022langchain}
enforces a character-count ceiling that the user must manually translate
into an approximate token count; LlamaIndex's \texttt{SentenceSplitter}
\cite{liu2022llamaindex} improves coherence but still requires a
user-supplied token limit; DrQA \cite{chen2017drqa} pre-segments Wikipedia
into fixed-length passages without model-specific adaptation. Chonkie \cite{chonkie2024} is a more recent lightweight, chunking-focused
library, but its splitters likewise take a user-supplied chunk size and, by
default, count in characters rather than model tokens (Section~\ref{sec:chunking-correctness}).
In each case,
chunking is an internal utility of a larger retrieval framework, or a standalone
splitter that still requires manual token configuration, rather than an
independently versioned, \emph{model-aware} resource. Token-aware splitting
itself is one line in these libraries once the user supplies the tokenizer; to
our knowledge, what no prior open library ships is a maintained
model-to-tokenizer-and-budget \emph{registry} that resolves those parameters
automatically from a named model and exposes the whole as a standalone package
(Table~\ref{tab:comparison}).

\begin{table}[t]
  \centering
  \small
  \resizebox{\columnwidth}{!}{%
  \begin{tabular}{lccccc}
    \toprule
    \textbf{Tool} & \textbf{Model-} & \textbf{Auto} & \textbf{Post-chunk} &
    \textbf{Light-} & \textbf{Async} \\
    & \textbf{aware} & \textbf{registry} & \textbf{ranking} &
    \textbf{weight} & \textbf{stream} \\
    \midrule
    LangChain splitter  & No  & No  & No  & No  & No  \\
    LlamaIndex splitter & No  & No  & No  & No  & No  \\
    Chonkie             & No  & No  & No  & Yes & No  \\
    \textbf{ChunkRank}  & \textbf{Yes} & \textbf{Yes} & \textbf{Yes} &
    \textbf{Yes} & \textbf{Yes} \\
    \bottomrule
  \end{tabular}}
  \caption{Feature comparison with the widely-used chunking utilities.
           ``Model-aware'' means chunk size is derived from the target model's
           tokenizer and context window without manual configuration
           (LangChain and Chonkie can be pointed at a tokenizer, but the user
           must supply it; neither resolves it automatically from a model name);
           ``Auto registry'' means the library ships pre-configured parameters
           for named models. Quantitative overflow/utilisation results are in
           Table~\ref{tab:overflow}.}
  \label{tab:comparison}
\end{table}

\subsection{Answer Ranking and Passage Re-ranking}

The classical BM25 function \cite{robertson2009bm25}, built on TF-IDF weighting
\cite{sparckjones1972idf}, remains a strong baseline \cite{thakur2021beir}.
Neural re-ranking with BERT \cite{devlin2019bert,nogueira2019passage} produced
monoT5 \cite{nogueira2020monot5}, ColBERT \cite{khattab2020colbert}, and SPLADE
\cite{formal2021splade}; sentence-level dense encoders \cite{reimers2019sbert}
now dominate semantic similarity and back most dense retrieval
\cite{karpukhin2020dpr}. All target \emph{retrieval}, finding the most relevant
passage in a large corpus, typically requiring a full retrieval stack. ChunkRank
instead ranks among candidate answers from the chunks of a single document, with
no corpus, index, or retrieval stack, a structurally simpler \emph{post-chunk}
setting not previously packaged as a standalone resource.

\subsection{Retrieval-Augmented Generation}

RAG \cite{lewis2020rag} combines a retriever with a generative model. Extensions
such as FiD \cite{izacard2021fid}, REPLUG \cite{shi2023replug}, Self-RAG
\cite{asai2024selfrag}, RECOMP \cite{xu2023recomp}, and GraphRAG
\cite{edge2024graphrag} couple retrieval, compression, or graph structure to a
specific generator, and are released, if at all, as research code rather than reusable
middleware. ChunkRank is not a RAG system (no index, no corpus, no modified
generation) but drops into any RAG pipeline. Nor do longer context windows (128k
in GPT-4 \cite{openai2023gpt4} to 10M in Llama 4 Scout \cite{meta2025llama4})
eliminate chunking: cost scales with length, long contexts degrade positional
attention \cite{liu2024lostmiddle}, and most self-hosted or edge models keep small
budgets, so a maintained model-aware chunker stays relevant as budgets diversify
rather than converge.

\section{Problem Definition}
\label{sec:problem}

Let $D$ be a document and $M$ a language model with tokenizer
$\mathcal{T}_M$, maximum context window $W_M$ (tokens), and a token budget
$R_M$ reserved for the prompt template and generated output. Define the
\emph{effective chunk budget} $B_M = W_M - R_M$.

\subsection{Model-Aware Chunking}

A chunking function $\mathcal{C}_M$ partitions $D$ into an ordered sequence
$\langle c_1, \ldots, c_n \rangle$ subject to the hard constraint
$\forall\, i: |\mathcal{T}_M(c_i)| \leq B_M$ and the soft objective of
preserving coherence across boundaries. A user-configurable overlap of
$\delta < B_M$ tokens ($\text{suffix}_{c_i}=\text{prefix}_{c_{i+1}}$) trades
storage for continuity.

\subsection{Post-Chunk Answer Selection}

Given a query $q$ and the answers $a_i$ produced by applying the model to each
chunk $c_i$ (with reader score $s_i$ where available), let
$A = \{(a_i, i, s_i) : a_i \neq \varnothing\}$ be the non-empty candidates. The
selection objective is:
\[
  a^* = \arg\max_{(a_i, i, s_i) \in A}\; \mathcal{S}(q, a_i, i, s_i)
\]
where $\mathcal{S}$ maps a candidate, its position $i$, and its score $s_i$ to a
real value. This general form covers content-based scorers that use only
$(q, a_i)$ (BM25, TF-IDF, embedding, cross-encoder), a position-based rule
($\texttt{first}$: $\mathcal{S} = -i$), and a score-based rule
($\texttt{confidence}$: $\mathcal{S} = s_i$). ChunkRank supports six
instantiations, detailed in Section~\ref{sec:ranker}.

Note that this formulation is \emph{extractive at the meta-level}: it
selects among already-generated answers rather than generating a new one,
which preserves the determinism of the underlying model call.

\section{System Design}
\label{sec:design}

\begin{figure}[h]
\centering
\resizebox{0.85\columnwidth}{!}{%
\begin{tikzpicture}[
  box/.style      = {rectangle, rounded corners=3pt, draw=black!70, fill=blue!8,
                     minimum width=28mm, minimum height=7mm,
                     font=\normalsize\sffamily, align=center},
  inputbox/.style = {rectangle, rounded corners=3pt, draw=black!50, fill=gray!10,
                     minimum width=28mm, minimum height=7mm,
                     font=\normalsize\sffamily, align=center},
  outputbox/.style= {rectangle, rounded corners=3pt, draw=green!50!black,
                     fill=green!8, minimum width=28mm, minimum height=7mm,
                     font=\normalsize\sffamily, align=center},
  arr/.style      = {-{Stealth[length=5pt]}, thick},
  darr/.style     = {-{Stealth[length=5pt]}, thick, dashed, gray},
]
\node[inputbox]                   (doc)   {Document $D$};
\node[inputbox, right=8mm of doc](model) {Model name $M$};
\node[box, below=5mm of model]   (reg)   {Model Registry};
\node[box, below=4mm  of reg]     (tok)   {Tokenizer Adapter};
\node[box, below=4mm  of tok]     (chunk) {Chunker};
\node[box, below=4mm  of chunk]   (llm)   {LLM calls (per chunk)};
\node[box, below=4mm  of llm]     (rank)  {Ranker};
\node[outputbox, below=4mm of rank](out)  {Best answer $a^*$};
\node[inputbox, right=6mm of rank](query){Query $q$};
\node[inputbox, right=6mm of chunk](cache){Chunk Cache};
\draw[arr] (doc)   -- (reg);
\draw[arr] (model) -- (reg);
\draw[arr] (reg)   -- (tok)   node[midway, right, font=\small\sffamily] {$W_M,\;\mathcal{T}_M$};
\draw[arr] (tok)   -- (chunk) node[midway, right, font=\small\sffamily] {$B_M$};
\draw[arr] (chunk) -- (llm)   node[midway, right, font=\small\sffamily] {$c_1\ldots c_n$};
\draw[arr] (llm)   -- (rank)  node[midway, right, font=\small\sffamily] {$a_1\ldots a_n$};
\draw[arr] (rank)  -- (out);
\draw[arr] (query.west) -- (rank.east);
\draw[darr] (chunk.east) -- (cache.west)
            node[midway, above, font=\small\sffamily] {miss};
\draw[darr] (cache.south) -- ++(0,-5mm) -| (llm.east)
            node[pos=0.3, below, font=\small\sffamily] {hit};
\end{tikzpicture}%
}
\caption{ChunkRank pipeline. Solid arrows show the primary data flow;
         dashed arrows show the optional disk-cache path.
         $W_M$ is the model context window, $\mathcal{T}_M$ its tokenizer,
         and $B_M = W_M - R_M$ the effective chunk budget.}
\label{fig:architecture}
\end{figure}
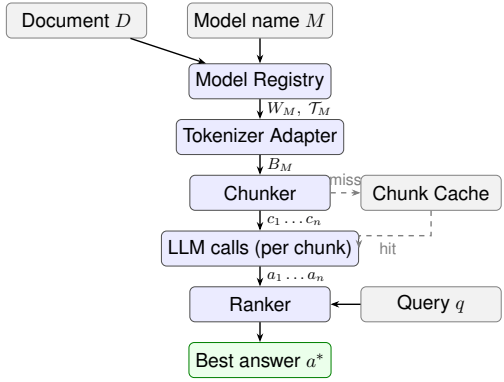

ChunkRank is organized into four components that run sequentially:
\emph{Model Registry} $\to$ \emph{Tokenizer Adapter} $\to$ \emph{Chunker}
$\to$ \emph{Ranker} (Figure~\ref{fig:architecture}, Appendix~\ref{app:arch}).
Each component exposes a clean interface and can be
replaced or extended independently, which is what allows the resource to be
adopted incrementally rather than as an all-or-nothing framework.

\subsection{Model Registry}
\label{sec:registry}

The registry is the single source of truth for model-specific parameters,
and is the primary reusable artifact this paper releases as a community
resource. Each entry stores five fields: the model \texttt{name},
\texttt{max\_context} (context window in tokens), the \texttt{tokenizer} backend
(\texttt{hf} or \texttt{tiktoken}), the \texttt{tokenizer\_id} (tokenizer name or
encoding), and \texttt{default\_reserve} (tokens held back for the prompt
template and generated output).

The registry ships with 90 pre-configured models across 15 providers:
OpenAI, Anthropic, Google, Meta, Mistral, Microsoft, Alibaba (Qwen),
Cohere, DeepSeek, TII, EleutherAI, IBM, xAI, AllenAI, and Hugging Face.
Context windows range from 512 tokens (BERT-base \cite{devlin2019bert}) to
10M tokens (Llama 4 Scout \cite{meta2025llama4}). The registry is stored as
a plain JSON file distributed with the package, so it can be inspected,
diffed, and extended by the community through ordinary version control
without touching library code.

Runtime model registration is supported via a one-line API call
(\texttt{chunkrank.register\_model(name, max\_context=..., tokenizer=...,
tokenizer\_id=..., default\_reserve=...)}), so the resource does not need
to be re-released every time a provider ships a new model. Runtime
entries are stored in memory and take precedence over the static JSON
registry, eliminating the need to upgrade the library when a new model
releases. If a model is absent from both sources, ChunkRank falls
back to a safe default: a 128k-token context window, tiktoken
\texttt{o200k\_base} encoding, and a 512-token reserve.

\paragraph{Registry validation.} To \emph{verify} the 90 entries rather than
trust them, we validate every entry along three axes and release the validation
harness with the library. (i)~\emph{Schema and consistency}: all 90 entries
type-check and satisfy $\text{max\_context} > \text{reserve} > 0$. (ii)~%
\emph{Tokenizer resolution}: all 55 tiktoken encodings resolve, and of the 35
Hugging Face entries, 19 load in a stock environment while the rest are gated
models that load for authenticated users or require a newer
\texttt{transformers}; no entry is malformed. 41 non-OpenAI models are mapped
to \texttt{o200k\_base} as an \emph{approximate} token counter (exact tokenizers
being proprietary or gated); on English this under-counts by $0.3$--$11.3\%$ vs.\
the true tokenizer where public (Appendix~\ref{app:proxy}), an overflow risk the
default \texttt{reserve} absorbs at typical budgets but not necessarily at very
large ones (Limitation~8), with a larger cross-script error quantified in
Section~\ref{sec:multilingual}. (iii)~\emph{Context-window audit}: we check
every Anthropic entry against the provider's live Models API and a documented
subset of 27 other-provider models against published specifications. The audit
caught two stale context windows, later-revised models listed at 200k that in
fact expose a 1M window, which we corrected, and the 27 documented values all
match. Shipping the harness lets the community re-run these checks as the
registry grows, so registry correctness is a maintained, testable property
rather than a one-time claim.

\subsection{Tokenizer Adapter}
\label{sec:tokenizer}

Tokenizer implementations differ substantially in API surface and
initialization overhead. The Tokenizer Adapter presents a uniform two-method
interface (\texttt{encode(text) $\to$ List[int]}, \texttt{count(text) $\to$
int}) to the rest of the system. Three backends are supported:

\begin{itemize}
  \item \textbf{tiktoken}: Used for OpenAI, Anthropic, Mistral, Cohere,
        DeepSeek, and Qwen models, with the encoding label loaded directly
        from the registry entry and the \texttt{tiktoken} object cached
        per process.

  \item \textbf{Hugging Face Transformers} \cite{wolf2020transformers}:
        Used for Meta Llama and encoder-based models (BERT, T5,
        Longformer, BigBird), loaded via
        \texttt{AutoTokenizer.\allowbreak from\_pretrained()} with fast tokenization
        enabled.

  \item \textbf{Character-ratio fallback}: When neither tiktoken nor
        Transformers is installed, token count is approximated as
        $\lfloor |s| / 4 \rfloor$, where $|s|$ is the UTF-8 character
        length, conservative, dependency-free, but less accurate for
        non-Latin scripts where the ratio can exceed 4:1.
\end{itemize}

Switching tokenizer backends requires only a registry update; no application
code changes are needed. This design choice is what keeps the resource
usable in environments where heavier tokenizer dependencies cannot be
installed.

\subsection{Chunker}
\label{sec:chunker}

The Chunker receives a document string and the model's effective token
budget $B_M$ from the adapter, and returns a list of text segments each
satisfying the hard token constraint. The budget is derived by default from the
model's context window (Section~\ref{sec:registry}) but remains a configurable
parameter. Because the accuracy-optimal budget is dataset-dependent, ChunkRank
exposes it rather than hard-coding a split size. Two strategies are available.

\paragraph{Token-Budget Sliding Window (default).}

This strategy operates without sentence boundary detection, making it
suitable for any language and requiring no additional dependencies
(Algorithm~\ref{alg:sliding}, Appendix~\ref{app:alg}).
Rather than shrinking the window by a fixed factor per iteration, the
inner loop binary-searches the character range $[\text{pos}+1,
\text{end}]$ (width at most $B \cdot r$), converging in $O(\log B)$
tokenizer calls per chunk since $r$ is a constant, tighter and more
predictable than a geometric-decay scheme as the effective budget $B$
grows for larger-context models. Overlap is applied in character space at
the same 4:1 ratio, preserving $\delta$ tokens of context without an extra
tokenizer call per chunk.

\paragraph{Semantic Similarity (optional).}

When \texttt{sentence-transformers} \cite{reimers2019sbert} is installed,
ChunkRank can group sentences by cosine similarity before enforcing the
token budget, opening a new chunk when the next sentence would violate it.
This benefits documents with distinct topical sections at the cost of a
one-time embedding pass.

\subsection{Ranker}
\label{sec:ranker}

The Ranker selects one answer from the per-chunk candidates. Empty answers
(chunks for which the reader or generator produced nothing) are dropped first,
so every method operates on the same non-empty candidate set. When a method
assigns equal scores to all candidates, ChunkRank breaks ties by \emph{document
order}; a method with no discriminative signal therefore degenerates gracefully
to first-chunk selection rather than to an arbitrary choice. Six methods are
available, with \texttt{first} as the default:

\paragraph{First non-empty (default).}

Returns the answer from the first chunk that produced a non-empty output
(document order). It performs no query--answer scoring and has no dependencies;
it is \emph{abstention-aware}, exploiting the reader's tendency to return nothing
on chunks that do not contain the answer. It is the default because, empirically
(Section~\ref{sec:experiments}), it matches or beats every content-based method
on our datasets.

\paragraph{Reader confidence.}

Selects the non-empty answer with the highest caller-supplied confidence (e.g.\
an extractive reader's span score or a generator's log-probability), the
classic multi-passage reading-comprehension selection rule \cite{chen2017drqa}.
Deterministic given the scores, with no additional dependencies.

\paragraph{BM25.}

BM25 \cite{robertson2009bm25} scores a document $d$ against query $q$ as:
\begin{align*}
  \text{BM25}(q, d) = \sum_{t \in q} \text{IDF}(t) \cdot {} \\
  \frac{f(t, d) \cdot (k_1 + 1)}{f(t, d) + k_1 \cdot \left(1 - b + b \cdot \frac{|d|}{\text{avgdl}}\right)}
\end{align*}
where $f(t, d)$ is term frequency in $d$, $\text{IDF}(t)$ is inverse
document frequency over the candidate set, and $k_1 = 1.5$, $b = 0.75$ are
standard parameters (\texttt{rank-bm25} implementation). BM25 is entirely
deterministic, requiring no floating-point precision choices, so reproducing a
result never depends on a hardware- or version-dependent forward pass.

\paragraph{TF-IDF Cosine Similarity.}

Query and answers are vectorized via scikit-learn's
\texttt{TfidfVectorizer} \cite{pedregosa2011sklearn,sparckjones1972idf};
similarity is the cosine between query and answer vectors, also
deterministic, but normalizing document length differently from BM25.

\paragraph{Dense Embedding.}

Query and answers are encoded by a sentence-level transformer into
fixed-length vectors $\mathbf{e}_q, \mathbf{e}_a \in \mathbb{R}^d$, scored
by cosine similarity \cite{reimers2019sbert}. This captures semantic
paraphrase invisible to lexical methods, but requires
sentence-transformers and is sensitive to encoder choice.

\paragraph{Cross-Encoder.}

A cross-encoder \cite{nogueira2019passage,wang2020minilm} scores the
concatenated pair $(q, a_i)$ directly (\texttt{ms-marco-MiniLM-L-6-v2} by
default); cross-encoders consistently outperform bi-encoders on
re-ranking \cite{thakur2021beir} but are slower, requiring a separate
forward pass per candidate.

Table~\ref{tab:ranker_comparison} (Appendix~\ref{app:ranker}) summarizes the
trade-offs: \texttt{first} and reader-confidence are dependency-free selection
rules (the default, \texttt{first}, performs no query--answer scoring at all);
BM25 and TF-IDF are deterministic and dependency-light but purely lexical, while
embedding and cross-encoder capture semantics at the cost of
\texttt{sentence-transformers} and slower inference.

\section{Implementation}
\label{sec:impl}

\subsection{Dependency Philosophy}

ChunkRank targets environments where installing heavy ML stacks is
impractical (CI systems, serverless functions, restricted clusters, edge
deployments); minimizing installation friction is a design requirement,
not an afterthought. The core system (registry, tiktoken and
character-fallback tokenizer adapters, sliding-window chunker, BM25 and
TF-IDF ranking) requires only three runtime dependencies: \texttt{numpy}
($\geq$1.26, array operations), \texttt{scikit-learn} ($\geq$1.5, TF-IDF
vectorization and cosine similarity), and \texttt{rank-bm25} ($\geq$0.2.2,
BM25 scoring).
Optional extras activate neural capabilities via
\texttt{pip install chunkrank[semantic]} (sentence-transformers) or
\texttt{chunkrank[all]} (all optional backends); PyTorch is never a hard
dependency and is only imported when the user selects
\texttt{method="embedding"} or \texttt{method="cross-encoder"}.

\subsection{API Design}

ChunkRank exposes three interaction levels for different integration effort: a
one-shot \textbf{Function API} for scripts and notebooks, a stateful
\textbf{Pipeline API} that caches chunker configuration across repeated queries
over one document, and a \textbf{Component API} exposing the chunker and ranker
directly; all three are shown in Appendix~\ref{app:api}.

\subsection{Async, Caching, and Compatibility}

\texttt{AsyncChunkRankPipeline} runs CPU-bound work in threads
(\texttt{asyncio.to\_thread()}) and LLM calls as coroutines so neither blocks
the event loop; both pipelines expose a \texttt{stream()} method. \texttt{ChunkCache}
persists chunked documents to disk keyed by a SHA-256 hash of (text, model,
strategy, overlap), skipping chunking on a hit, using only the standard library.
ChunkRank targets Python 3.10--3.13, uses \texttt{importlib.resources.files()}
(PEP 451) for registry access, and its test suite passes on all four versions.

\section{Resource Release and Reproducibility}
\label{sec:resource}

Beyond the library, we release the artifacts needed to reproduce and extend
this work, documenting what is available, how it is versioned, and how it is
maintained.

\paragraph{Distribution, licensing, and versioning.} ChunkRank is
published on PyPI as the \texttt{chunkrank} package and on GitHub at
\url{https://github.com/AmitoVrito/chunkrank} under the Apache 2.0
license, permitting unrestricted commercial and academic reuse, including
inside larger RAG frameworks. The repository hosts the source code, the
versioned model registry JSON (Section~\ref{sec:registry}, diffable
independently of code to audit exactly which models changed between
releases), and the public test suite. The package follows semantic
versioning; the results in this paper use release \texttt{2.0.0}. As of 2026-09-23, it has been downloaded 8{,}806
times from PyPI (554 in the preceding 30 days), a figure that includes mirror
and CI traffic and is therefore an upper bound, but one that indicates adoption
beyond the authors' own usage.

\paragraph{Test coverage and evaluation artifacts.} The public test suite
exercises the registry, tokenizer adapters, both chunking strategies, all
six answer-selection methods, the disk cache, and the sync and async pipelines,
and runs on CPython 3.10 through 3.13. Alongside the library, we release
the evaluation harness used in Section~\ref{sec:experiments}: the dataset
filtering script, the evaluation configuration, and the full per-method
result tables (Tables~\ref{tab:results}, \ref{tab:hotpot}, and
\ref{tab:latency}), so the empirical claims here can be independently
checked and extended to new datasets or selection methods without
re-implementing the pipeline.

\paragraph{Computational reproducibility.} The reported numbers were produced
with \texttt{chunkrank} \texttt{2.0.0} and pinned dependencies
(\texttt{transformers}~4.44.2, \texttt{sentence-transformers}~3.0.1,
\texttt{torch}~2.2.2 on CPU, \texttt{numpy}$<$2). To make neural ranking
portable across machines, every model is pinned to an exact Hugging Face
revision (reader \texttt{deepset/\allowbreak roberta-\allowbreak base-\allowbreak squad2}@\allowbreak\texttt{adc3b06f}, embedding
\texttt{all-\allowbreak MiniLM-\allowbreak L6-\allowbreak v2}@\allowbreak\texttt{1110a243}, cross-encoder
\texttt{ms-\allowbreak marco-\allowbreak MiniLM-\allowbreak L-6-\allowbreak v2}@\allowbreak\texttt{233902d2}) and run with
\texttt{device="cpu"}; we verified the neural rankings are bit-identical on CPU
and Apple MPS. The released evaluation harness invokes \texttt{chunkrank.Ranker}
directly for every selection method, so a re-run reproduces the paper's numbers
from the installed package rather than a separate re-implementation.

\paragraph{Extensibility.} Three extension points support community
contribution without forking the library: new registry entries as a JSON
diff, new chunking strategies behind the existing \texttt{Chunker}
interface, and new ranking methods behind the existing \texttt{Ranker}
interface, the same interfaces used internally, so third-party
extensions are not second-class citizens.

\section{Experiments}
\label{sec:experiments}

\subsection{Task and Motivation}

We evaluate ChunkRank on \emph{long-document single-hop QA}: each example
pairs a document exceeding the model's chunk budget with a factual question
whose answer lies in one contiguous passage, and candidate answers come from
querying the QA model (extractive reader or generative LLM) against each chunk
independently. This validates real-world
utility rather than a leaderboard, and we release the harness
(Section~\ref{sec:resource}) for reuse. Unlike RAG/re-ranking benchmarks (BEIR
\cite{thakur2021beir}, MS MARCO \cite{bajaj2016msmarco}) that search a
\emph{corpus} of thousands of passages, our candidate set is small (typically
3--20 chunks from one document) and pre-defined, the dominant case for
practitioners applying LLMs to individual documents, yet one with no
dedicated benchmark or reusable harness prior to this release.

\subsection{Experimental Setup}

\paragraph{Data.}

We build evaluation sets from NaturalQuestions (NQ)
\cite{kwiatkowski2019naturalquestions} and TriviaQA \cite{joshi2017triviaqa},
keeping examples whose answer context exceeds 8{,}000 tokens (forcing chunking
at a 6{,}000-token budget): \textbf{500} examples each.

\paragraph{Answer generation and chunking.} Candidate answers are extracted per
chunk with \texttt{deepset/roberta-base-squad2} \cite{devlin2019bert}, an
extractive RoBERTa QA model \cite{liu2019roberta}, the standard paradigm for
NQ/TriviaQA, fully reproducible, CPU-only, no fine-tuning, using ChunkRank's
sliding window (6{,}000-token budget, 64-token overlap; mean 3.5 / 4.0 chunks
per document on NQ / TriviaQA). Because the extractive reader has a 512-token input
limit, each chunk is processed by the Hugging~Face question-answering pipeline,
which slides a 512-token window (stride 128) over the \emph{whole} chunk and returns
its single highest-confidence span (or nothing, when it finds no answer). The reader
therefore both extracts an answer per chunk and abstains on chunks that lack one;
the per-chunk confidence it produces is what the \texttt{confidence} selector uses,
while the content-based methods (Section~\ref{sec:ranker}) re-score the returned
answer \emph{strings} against the query. On a chunk that contains no answer the
reader usually returns nothing, so a large fraction of chunks are empty (52\% on NQ,
57\% on TriviaQA for this reader; Table~\ref{tab:empty}), which is central to the
results below. Section~\ref{sec:generative} repeats the study with a generative
reader that leaves far fewer examples with no answer at all.

\paragraph{Baselines and metrics.} All methods select over the non-empty
candidates. \textbf{First (non-empty)} / \textbf{Last} return the first / last
non-empty candidate; \textbf{Random} samples one uniformly; \textbf{Oracle} takes
the candidate with the highest F1 against the gold answer, the best any selector
could achieve, an upper bound. We report Exact Match (EM) and token-level F1
\cite{kwiatkowski2019naturalquestions} with $95\%$ bootstrap confidence intervals
and paired bootstrap tests against first-non-empty.

\begin{table}[t]
  \centering
  \small
  \resizebox{\columnwidth}{!}{%
  \begin{tabular}{lcc}
    \toprule
    \textbf{Reader / dataset} & \textbf{Empty chunks} & \textbf{No-answer ex.} \\
    \midrule
    NQ, extractive        & 52\% & 22\% \\
    TriviaQA, extractive  & 57\% & 16\% \\
    HotpotQA, extractive  & 85\% & 38\% \\
    \midrule
    NQ, generative        & 36\% &  2\% \\
    TriviaQA, generative  & 51\% &  7\% \\
    HotpotQA, generative  & 79\% & 18\% \\
    \bottomrule
  \end{tabular}}
  \caption{Abstention rates. ``Empty chunks'' is the share of chunks for which the
           reader returned no answer; ``No-answer ex.'' is the share of examples
           with no non-empty candidate at all (every method scores 0). The
           extractive reader reads full chunks. These rates drive the results:
           first-non-empty exploits exactly this abstention.}
  \label{tab:empty}
\end{table}

\subsection{Chunking Correctness}
\label{sec:chunking-correctness}

Before evaluating ranking, we validate the resource's \emph{primary}
contribution: chunking that never exceeds the target model's budget. We split the
500 long TriviaQA documents for \texttt{gpt-4o-mini} (6{,}000-token budget,
\texttt{o200k\_base}) with ChunkRank, LangChain's
\texttt{RecursiveCharacterTextSplitter} \cite{chase2022langchain}, and Chonkie's
\texttt{TokenChunker} \cite{chonkie2024}, counting chunks over 6{,}000 tokens and
budget utilisation (Table~\ref{tab:overflow}). Character-based splitting, the
default in both LangChain (a character ceiling) and Chonkie (a character backend),
overflows $0.6\%$ of chunks, the largest reaching 7{,}397 tokens ($23\%$ over
budget), because the character-to-token ratio varies across documents. Shrinking
the ceiling until overflow vanishes (LangChain 12k) wastes $61\%$ of the budget
and yields $1.9\times$ more chunks. Token-aware \emph{configurations} of both
libraries reach zero overflow at high utilisation, but only once the user
manually supplies the target model's tokenizer. ChunkRank reaches zero overflow
\emph{automatically} from the model name, at lower utilisation ($75.9\%$ vs.\
$84$--$86\%$), a deliberate safety margin. Its zero overflow is guaranteed by
construction (it splits with the same tokenizer it is measured against, which is
the point: that tokenizer is the target model's). The contribution is thus not
token counting (any library can be configured for it) but deriving tokenizer
and budget from a named model without manual setup.

\begin{table}[t]
  \centering
  \small
  \resizebox{\columnwidth}{!}{%
  \begin{tabular}{lccc}
    \toprule
    \textbf{Splitter} & \textbf{Overflow} & \textbf{Mean util.} & \textbf{Max tok.} \\
    \midrule
    LangChain char (naive 4:1)      & 0.64\% & 75.0\% & 7{,}375 \\
    Chonkie char (default)          & 0.60\% & 76.0\% & 7{,}397 \\
    LangChain char (aggressive)     & 0.00\% & 39.3\% & 4{,}559 \\
    LangChain tiktoken (manual)     & 0.00\% & 84.5\% & 5{,}979 \\
    Chonkie tiktoken (manual)       & 0.00\% & 86.2\% & 6{,}000 \\
    \textbf{ChunkRank (auto)}       & \textbf{0.00\%} & 75.9\% & \textbf{6{,}000} \\
    \bottomrule
  \end{tabular}}
  \caption{Chunk overflow rate, mean budget utilisation, and largest chunk
           across 500 TriviaQA documents (\texttt{gpt-4o-mini}, 6{,}000-token
           budget). Character-based splitters (top two) overflow the budget;
           token-aware ones (middle) do not, but require manual per-model
           tokenizer configuration. ChunkRank guarantees zero overflow
           automatically from the model name, at lower utilisation
           ($75.9\%$ vs.\ $84$--$86\%$), a deliberate safety margin.}
  \label{tab:overflow}
\end{table}

\subsection{Multilingual Tokenizer Fidelity}
\label{sec:multilingual}

Our QA evaluation is English-only, but the chunking contribution can be probed
multilingually. The character-ratio fallback ($\lfloor|s|/4\rfloor$) is calibrated
for Latin-script English; measured on UDHR Article~1 across 11 languages
(Appendix~\ref{app:multi}), it over-counts English by $27\%$ (safe) but
\emph{under}-counts logographic and syllabic scripts by up to $71\%$ (Chinese,
Japanese), since one character maps to roughly one token. Consequently a
character window safe at 6{,}000 English tokens holds $\sim$20{,}000 tokens of
Japanese ($3.4\times$ budget) and overflows for \emph{every} non-Latin script
tested. Only a tokenizer-derived budget, which ChunkRank computes from the model
name, is safe across scripts (hence the fallback is opt-out, not default).

\subsection{Answer Selection Results}

Table~\ref{tab:results} reports EM and F1 at the 6{,}000-token budget with $95\%$
bootstrap CIs. Every method selects over one candidate set per example: the
non-empty answers the reader produced, with empty spans dropped as the library
does (Section~\ref{sec:ranker}). The headline is a negative result. On neither NQ
nor TriviaQA does any content-based ranker (lexical, dense-embedding,
cross-encoder) significantly beat simply taking the first non-empty answer:
embedding ties first-non-empty or falls below it, while reader-confidence, the
other reader-based rule, ties it (and is numerically higher on TriviaQA,
$p=0.18$). The oracle headroom that remains is captured by no method we tested.
BM25 and TF-IDF add no signal on short factoid answers, which rarely share query
terms, so all their candidate scores are equal on $79$--$90\%$ of multi-candidate
examples and they fall back to document order, reducing in practice to first-chunk.

\paragraph{Abstention, not ranking, does the work.} First-non-empty is hard to beat
because the reader abstains: on a chunk that does not contain the answer it returns
nothing, so the first non-empty answer is the answer from the first chunk the
reader was confident about. Re-scoring the returned answer \emph{strings} against
the query adds little on top of this implicit filtering: real oracle headroom
remains (Table~\ref{tab:results}), but no content-based selector captures it, and
the reader-based rules (first-non-empty, reader-confidence) are the strongest we
tested.

\paragraph{Where there is a genuine choice.} A tie is trivial when an example has
only one non-empty candidate, so we restrict to examples with $\geq 2$ non-empty
candidates, where a selector must actually choose. With the generative reader this
is the common case ($n{=}70/100$ on NQ, $58/100$ on TriviaQA, $181/400$ on
HotpotQA), and there first-non-empty is not merely tied but significantly
\emph{beats} every content-based method: 40.0 EM versus 22.9 (embedding) and 18.6
(cross-encoder) on NQ ($p<0.001$), 58.6 versus 44.8 on TriviaQA ($p=0.004$), and
38.7 versus 26.5 on HotpotQA ($p=0.002$). With the extractive reader reading full
chunks, roughly half of all examples now have a real choice ($n{=}242$ on NQ,
$247$ on TriviaQA, $122$ on HotpotQA), and there too no content-based method
significantly beats first-non-empty; on NQ it is significantly better (first-non-empty
27.7 vs.\ embedding 20.2, $p=0.004$), and on TriviaQA the differences are
non-significant.
The negative result is therefore not an artifact of single-candidate ties: where a
real choice exists, taking the first non-empty answer is at least as good as, and
under the stronger reader clearly better than, re-scoring the candidates.

\paragraph{The effect is not positional.} A tempting explanation for a strong
first-chunk baseline is positional bias in Wikipedia-derived data, where the answer
tends to sit near the start. We rule this out on HotpotQA (distractor)
\cite{yang2018hotpotqa}, where the answer-bearing paragraph sits at a near-uniform
position (mean index 4.6; only $10\%$ first; Appendix~\ref{app:hotpot}). Even there,
first-non-empty is not beaten by any content-based ranker (Table~\ref{tab:hotpot}).
It is abstention, not answer position, that makes first-non-empty strong, which is
why ChunkRank ships it as the default and exposes the content-based methods for
callers whose readers do not abstain.

\begin{table}[t]
  \centering
  \small
  \begin{tabular}{lcccc}
    \toprule
    \multirow{2}{*}{\textbf{Method}} &
    \multicolumn{2}{c}{\textbf{NaturalQuestions}} &
    \multicolumn{2}{c}{\textbf{TriviaQA}} \\
    \cmidrule(lr){2-3} \cmidrule(lr){4-5}
    & EM & F1 & EM & F1 \\
    \midrule
    First (non-empty) & \textbf{22.4} & \textbf{29.1} & 45.4 & 51.0 \\
    Last              & 14.0 & 20.7 & 40.8 & 47.1 \\
    Random            & 18.8 & 24.8 & 43.8 & 49.4 \\
    \midrule
    BM25              & 21.8 & 28.6 & 45.2 & 50.5 \\
    TF-IDF            & 20.4 & 27.4 & 44.8 & 51.2 \\
    Embedding         & 18.8 & 25.9 & 45.2 & 52.3 \\
    Cross-Encoder     & 19.6 & 27.0 & 45.2 & 51.7 \\
    Reader-conf.      & 20.8 & 27.0 & \textbf{47.4} & \textbf{53.2} \\
    \midrule
    Oracle            & 25.6 & 34.2 & 56.2 & 61.5 \\
    \bottomrule
  \end{tabular}
  \caption{Exact Match (EM) and token-level F1 for answer selection over the
           non-empty candidate set (chunk budget = 6{,}000 tokens, 500 examples
           per dataset; extractive reader reading full chunks). \textbf{Bold} marks
           the best non-oracle result per column. No content-based method (BM25,
           TF-IDF, embedding, cross-encoder) significantly beats first-non-empty: on
           NQ, embedding and cross-encoder are significantly \emph{below} it
           ($p<0.02$), and on TriviaQA all differences from first-non-empty are
           non-significant (reader-confidence is numerically highest, $p=0.18$). The
           reader-based selectors (first-non-empty, reader-confidence) match or beat
           the content-based ones throughout. Oracle headroom remains but is
           captured by no method.}
  \label{tab:results}
\end{table}

\subsection{Validation with a Generative LLM}
\label{sec:generative}

To test whether the finding depends on the extractive reader, we repeat the study
on 100 examples per dataset with a production \emph{generative} reader, Claude
Haiku 4.5 \cite{anthropic2025haiku}, queried through its API, with paired bootstrap
tests. Absolute scores rise sharply (the generative model is a far stronger
reader), but the conclusion holds: first-non-empty is not beaten. On NQ,
first-non-empty reaches 36.0 EM and embedding is significantly lower (24.0,
$p<0.001$); on TriviaQA, first-non-empty reaches 59.0 and embedding is again lower
(51.0, $p=0.006$). Crucially, the generative reader leaves far fewer examples with
no answer at all (2--18\% vs.\ 16--38\% for the extractive reader;
Table~\ref{tab:empty}), yet first-non-empty still wins, so the result is not an
artifact of a reader that abstains too readily. The oracle remains well above every
method (68.0 EM on TriviaQA), so real headroom exists that none of the tested
selectors captures. This run is at $n{=}100$ and released through the same harness,
which exposes both an extractive and a generative backend.

\subsection{Latency Analysis}

We benchmark chunking and selection latency on CPU (median of 100 runs; full table
in Appendix~\ref{app:latency}). Chunking a document into up to 11 chunks completes
in under 15\,ms, regardless of tokenizer backend. Selection is model-independent and
ranges from effectively free (\texttt{first} and \texttt{confidence}, the default,
and BM25 at $<0.2$\,ms) to $\sim$30\,ms for the neural methods at 5--20 candidates.
The whole pipeline adds under 50\,ms, negligible next to the seconds-scale reader
calls it wraps.

\subsection{When Not to Chunk: A Long-Context Baseline}
\label{sec:whennot}

Why chunk at all when a large-window model could read the whole document in one
call? We test this directly: for the same generative examples (NQ / TriviaQA
$n{=}100$, HotpotQA $n{=}400$), we send each full document to Claude Haiku 4.5 in
a single call (all documents fit its 200k-token window) and compare against
chunked first-non-empty selection (Table~\ref{tab:longcontext}).

On single-hop QA where the document fits, the two are statistically
indistinguishable (NQ 36.0 vs.\ 35.0, $p{=}0.87$; TriviaQA 59.0 vs.\ 57.0,
$p{=}0.70$), and total input tokens are essentially equal (chunked $1.02\times$
single-call, since the chunks reconstruct the document plus a small overlap and a
per-chunk prompt). For documents that fit a large window, chunking therefore
yields no accuracy or cost advantage. On \emph{multi-hop} HotpotQA, single-call
long-context wins (55.1 vs.\ 34.6 EM on the $n{=}136$ subset completed before an
API limit; the chunked score on that subset matches its full-set value of $33.5$,
confirming the subset is representative), because splitting a document into
256-token pieces separates the evidence a multi-hop question must combine. This is
the single-document, single-hop scope described in Limitation~1. We include HotpotQA only as a
probe of that limitation: its documents average $\sim$1.3k tokens and were
deliberately split at a 256-token budget, so this is not evidence that
long-context beats chunking in general.

This defines when the resource applies: chunking gives no benefit when a document
fits a large-window model, but it is \emph{required} for the many registry models
with 512--32k-token windows (self-hosted and edge models) and for documents beyond
any context window, where a single call is impossible.

\begin{table}[t]
  \centering
  \small
  \resizebox{\columnwidth}{!}{%
  \begin{tabular}{lccc}
    \toprule
    \textbf{Method} & \textbf{NQ} & \textbf{TriviaQA} & \textbf{HotpotQA} \\
    \midrule
    Chunked (first-non-empty) & 36.0 & 59.0 & 34.6 \\
    Long-context (single call) & 35.0 & 57.0 & \textbf{55.1} \\
    \midrule
    $p$ vs.\ chunked & 0.87 & 0.70 & $<$0.001 \\
    \bottomrule
  \end{tabular}}
  \caption{Whole-document single-call (Claude Haiku 4.5) vs.\ chunked
           first-non-empty, EM. Single-hop NQ/TriviaQA ($n{=}100$): a tie, at
           essentially equal input-token cost. Multi-hop HotpotQA
           ($n{=}136$ subset; chunked matches its full-set $33.5$): long-context
           wins, as expected when chunking splits multi-hop evidence
           (Limitation~1). All documents fit the 200k-token window.}
  \label{tab:longcontext}
\end{table}

\section{Conclusion}
\label{sec:conclusion}

We have presented ChunkRank, an open-source, versioned resource that packages
model-aware text chunking and post-chunk answer ranking as lightweight, modular
middleware for LLM pipelines. Unlike chunking utilities bundled inside larger
retrieval frameworks, it is a standalone, independently documented package: a
registry of 90 models across 15 providers, a three-tier API, a PyTorch-free
core, and only three dependencies, adoptable incrementally and extensible
without forking. All evaluation code, configuration, and results reported here are
released alongside it to support replication.

Experiments validate the chunking contribution and report a clear negative result
for answer selection. For chunking, ChunkRank produces zero budget overflow
automatically from the model name, where character-based splitters overflow or
waste the budget (Sections~\ref{sec:chunking-correctness}--\ref{sec:multilingual}).
For answer selection, across NaturalQuestions, TriviaQA, and HotpotQA, with an
extractive and a generative reader and paired bootstrap tests, no content-based
ranker (lexical, embedding, cross-encoder, or reader-confidence) reliably beats
taking the first non-empty answer: selection over chunks is dominated by the
reader's ability to abstain, not by re-scoring answer strings. A long-context
baseline further shows that chunking is not needed when a document fits a
large-window model, delimiting where the resource applies. The contribution is
thus not a stronger re-ranker but a modular, reproducible, community-maintainable
resource, with an honest account of what answer selection over chunks can and
cannot do. Future work includes multilingual calibration data, learned chunk
boundaries, multi-document fusion, and CPU-only embedding backends.

\section{Limitations}
\label{sec:limitations}

\begin{enumerate}

  \item \textbf{Single-document, single-hop scope.} ChunkRank processes one
        document at a time and post-chunk selection cannot synthesise evidence
        across chunks, so multi-document and multi-hop question answering are
        outside its current scope. Our HotpotQA study
        (Appendix~\ref{app:hotpot}) does not evaluate multi-hop reasoning: it
        concatenates the distractor paragraphs into a \emph{single} document and
        uses it only to test whether first-non-empty's strength is merely
        positional (it is not), with single-span answers, which is
        also why absolute scores there are capped.

  \item \textbf{Extractive answer selection.} The ranker selects among
        pre-generated answers. It does not merge complementary information
        from multiple chunks, a capability studied in FiD
        \cite{izacard2021fid}. If the answer to a query spans multiple
        chunks, ChunkRank may return an incomplete response.

  \item \textbf{Rule-based chunking.} Both chunking strategies apply fixed
        rules. We have not explored learning-based approaches that optimize
        chunk boundaries for downstream task performance.

  \item \textbf{Character-ratio fallback accuracy.} The len/$4$
        approximation is inaccurate for languages with high
        character-to-token ratios (e.g., Chinese, Japanese, Korean), where a
        single character may correspond to fewer than one BPE token. Users
        processing such texts should install either tiktoken or the Hugging
        Face tokenizer for the target model. As a released resource, this is
        a documented rather than silent limitation, and is a priority area
        for community-contributed multilingual calibration data in future
        releases.

  \item \textbf{Semantic chunking dependency.} Enabling the semantic
        similarity strategy requires \texttt{sentence-transformers}, which
        reintroduces a PyTorch dependency. This is at tension with the
        lightweight design goal; we intend to support lightweight CPU-only
        embedding backends in a future release.

  \item \textbf{Non-factoid queries.} The evaluation focuses on factoid QA
        where a single chunk contains the complete answer. For
        summarization, multi-hop reasoning, or open-ended generation, the
        ranking criteria and evaluation metrics require reconsideration.
        Post-chunk ranking as implemented here is not suited to queries
        whose correct answer requires synthesizing information across
        chunks.

  \item \textbf{Evaluation scope.} The answer-selection evaluation uses 500
        examples per dataset on NQ and TriviaQA and 400 on HotpotQA, with paired
        bootstrap confidence intervals and significance tests
        (Section~\ref{sec:experiments}), across three datasets spanning
        front-loaded and distributed answer positions, and we confirm the main
        finding with a generative reader in addition to extractive QA
        (Section~\ref{sec:generative}); the generative run is preliminary
        ($n{=}100$ per dataset). Three gaps remain: all three datasets are
        Wikipedia-derived, so domain diversity (legal, biomedical, code) is
        untested; the generative validation covers a single provider model; and our
        negative result concerns the specific selectors we implemented, so a
        learned or reader-aware selector could still beat first-non-empty. We
        release the evaluation harness, with extractive and generative backends, so
        the community can extend it to new domains, generators, and selectors
        without re-implementation.

  \item \textbf{English-centric fallback and proxy tokenizers.} The
        character-ratio fallback ($\lfloor |s| / 4 \rfloor$) is calibrated for
        Latin-script English; Section~\ref{sec:multilingual} quantifies its
        error across 11 languages (up to a $71\%$ token under-count for
        Chinese/Japanese, i.e.\ overflow risk), so it is an opt-out path, and
        the tiktoken and Hugging Face backends, which count tokens directly,
        are the correct choice for non-Latin scripts. Relatedly, 41 non-OpenAI
        models are mapped to \texttt{o200k\_base} as an \emph{approximate} token
        counter rather than their exact tokenizer; this under-counts by
        $0.3$--$11.3\%$ on English (Appendix~\ref{app:proxy}). The default
        per-model \texttt{reserve} absorbs this at typical budgets, but a fixed
        reserve does not scale: at a 128k-token budget an $11.3\%$ under-count is
        $\sim$14k tokens, far exceeding a 512--1{,}024-token reserve, so the
        zero-overflow guarantee is exact only for the 49 entries with a native
        tokenizer (tiktoken or Hugging~Face) and approximate for the 41
        proxy-mapped models. For those, or at very large budgets, the exact
        tiktoken or Hugging~Face tokenizer should be used. The QA evaluation
        itself remains English-only; multilingual QA is future work.

\end{enumerate}

\section*{Ethics Statement}

ChunkRank is infrastructure: it operates purely on documents and queries
passed to it by the calling application and does not itself collect,
transmit, or retain any user data. It maintains no server, no telemetry,
and no persistent state beyond the optional, entirely local, user-controlled
disk cache described in Section~\ref{sec:impl}, which stores only chunked
document text under a path the user chooses. ChunkRank does not call any
external service on its own behalf; the only network access in a typical
deployment is the LLM API call the user's own application makes, which is
entirely outside ChunkRank's control. Because ChunkRank is model- and
provider-agnostic infrastructure rather than a trained model, it does not
introduce new biases beyond those already present in whichever LLM and
tokenizer the user configures it to work with; it does not amplify or
mitigate those biases. The evaluation in Section~\ref{sec:experiments} uses
NaturalQuestions and TriviaQA, both public, widely used, English-language
QA benchmarks derived from Wikipedia and licensed for research use, and
introduces no new human-subjects data collection.

\section*{Acknowledgements}

The authors thank the open-source communities behind \texttt{rank-bm25},
\texttt{scikit-learn}, and \texttt{sentence-transformers} for maintaining
the libraries on which ChunkRank depends.

\bibliography{references}

\appendix

\section{Architecture Diagram}
\label{app:arch}

\section{Three-Tier API}
\label{app:api}

The Function, Pipeline, and Component APIs (Section~\ref{sec:impl}):

\nolinenumbers
\begin{lstlisting}
import chunkrank
# Level 1: one-shot function API
chunks = chunkrank.split(text, model="gpt-4o-mini")
answers = chunkrank.answer(question, chunks)  # calls your LLM
best = chunkrank.rank(answers)
# Level 2: stateful pipeline API
from chunkrank import ChunkRankPipeline
pipe = ChunkRankPipeline(model="gpt-4o-mini", retrieval_top_k=3)
answer = pipe.process(question, text)
# Level 3: component API
from chunkrank import Chunker, ChunkerConfig, Ranker
config = ChunkerConfig(model="claude-3-5-sonnet",
    strategy="semantic", overlap_tokens=64)
chunker, ranker = Chunker(config), Ranker(method="cross-encoder")
chunks = chunker.split(text)
answers = [my_llm(question, c) for c in chunks]  # your LLM call
ranked = ranker.rank(question, answers)
\end{lstlisting}
\linenumbers

\section{Token-Budget Sliding Window}
\label{app:alg}

Algorithm~\ref{alg:sliding} gives the default chunker (Section~\ref{sec:chunker}).
The inner loop binary-searches the character range $[\text{pos}+1,\text{end}]$
(width at most $B\cdot r$), converging in $O(\log B)$ tokenizer calls per chunk.

\begin{algorithm}[h]
  \caption{Token-Budget Sliding Window}
  \label{alg:sliding}
  \begin{algorithmic}[1]
    \REQUIRE document $D$, budget $B$, overlap $\delta$, initial ratio $r = 4$
    \STATE $\text{pos} \leftarrow 0$; \; $\text{chunks} \leftarrow []$
    \STATE $\text{approx} \leftarrow \max(64,\, B \cdot r)$ \hfill \COMMENT{character estimate}
    \WHILE{$\text{pos} < |D|$}
      \STATE $\text{end} \leftarrow \min(|D|,\, \text{pos} + \text{approx})$
      \IF{$|\mathcal{T}(D[\text{pos}:\text{end}])| > B$}
        \STATE $\text{lo}, \text{hi} \leftarrow \text{pos} + 1,\; \text{end}$ \hfill \COMMENT{binary search on chunk length}
        \WHILE{$\text{lo} < \text{hi}$}
          \STATE $\text{mid} \leftarrow \lceil (\text{lo} + \text{hi}) / 2 \rceil$
          \IF{$|\mathcal{T}(D[\text{pos}:\text{mid}])| \leq B$}
            \STATE $\text{lo} \leftarrow \text{mid}$
          \ELSE
            \STATE $\text{hi} \leftarrow \text{mid} - 1$
          \ENDIF
        \ENDWHILE
        \STATE $\text{end} \leftarrow \max(\text{pos} + 1,\, \text{lo})$
      \ENDIF
      \STATE $\text{chunks.append}(D[\text{pos}:\text{end}])$
      \IF{$\text{end} \geq |D|$}
        \STATE \textbf{break}
      \ENDIF
      \STATE $\text{pos} \leftarrow \text{end} - \delta \cdot r$ \textbf{if} $\delta > 0$ \textbf{else} $\text{end}$ \hfill \COMMENT{apply overlap}
    \ENDWHILE
    \RETURN chunks
  \end{algorithmic}
\end{algorithm}

\section{Multilingual Tokenizer Fidelity}
\label{app:multi}

Full per-language results for Section~\ref{sec:multilingual}
(Table~\ref{tab:multilingual}).

\begin{table}[h]
  \centering
  \small
  \begin{tabular}{llcc}
    \toprule
    \textbf{Language} & \textbf{Script} & \textbf{ch/tok} & \textbf{\texttt{len/4} err.} \\
    \midrule
    English  & Latin      & 5.15 & $+27\%$ \\
    Spanish  & Latin      & 4.50 & $+11\%$ \\
    French   & Latin      & 4.54 & $+12\%$ \\
    German   & Latin      & 4.32 & $+8\%$ \\
    Russian  & Cyrillic   & 3.81 & $-5\%$ \\
    Hindi    & Devanagari & 3.46 & $-15\%$ \\
    Arabic   & Arabic     & 2.70 & $-33\%$ \\
    Thai     & Thai       & 2.08 & $-49\%$ \\
    Korean   & Hangul     & 1.71 & $-59\%$ \\
    Chinese  & Han        & 1.23 & $-71\%$ \\
    Japanese & Japanese   & 1.18 & $-71\%$ \\
    \bottomrule
  \end{tabular}
  \caption{Characters per \texttt{o200k\_base} token and the error of the
           \texttt{len/4} character-ratio estimate on UDHR Article~1
           (negative error $=$ under-count $=$ overflow risk). Latin scripts are
           safe; logographic and abugida scripts under-count by up to $71\%$, so
           a character budget calibrated for English overflows the token budget.}
  \label{tab:multilingual}
\end{table}

\section{Proxy-Tokenizer Error}
\label{app:proxy}

The registry maps 41 non-OpenAI models to \texttt{o200k\_base} as an approximate
token counter. Table~\ref{tab:proxy} measures the resulting error on a fixed
8{,}000-character English document (o200k\_base: 1{,}513 tokens) against the true
tokenizer, for the non-OpenAI models whose tokenizer is public. The proxy
consistently \emph{under}-counts (the true model emits more tokens), by up to
$11.3\%$ (Mistral); a proxy under-count is an overflow risk, which the default
per-model \texttt{reserve} absorbs. Models with proprietary tokenizers (Claude,
Gemini, Grok, Cohere) cannot be measured directly.

\begin{table}[h]
  \centering
  \small
  \resizebox{\columnwidth}{!}{%
  \begin{tabular}{llcc}
    \toprule
    \textbf{Registry model} & \textbf{True tokenizer} & \textbf{True tok.} & \textbf{Proxy err.} \\
    \midrule
    qwen2.5-*    & Qwen2.5-7B        & 1{,}549 & $-2.3\%$ \\
    deepseek-v3  & deepseek-llm-7b   & 1{,}561 & $-3.1\%$ \\
    mistral-*    & Mistral-7B-v0.2   & 1{,}705 & $-11.3\%$ \\
    (older BPE)  & gpt2              & 1{,}517 & $-0.3\%$ \\
    \bottomrule
  \end{tabular}}
  \caption{Token count of the \texttt{o200k\_base} proxy (1{,}513) vs.\ the true
           tokenizer on an 8{,}000-character English document. Negative error $=$
           proxy under-count $=$ the true model emits more tokens than budgeted
           (overflow risk, absorbed by the reserve).}
  \label{tab:proxy}
\end{table}

\section{Ranker Method Trade-offs}
\label{app:ranker}

\begin{table}[h]
  \centering
  \small
  \resizebox{\columnwidth}{!}{%
  \begin{tabular}{lcccc}
    \toprule
    \textbf{Method}  & \textbf{Deterministic} & \textbf{Dependencies} &
    \textbf{Semantic} & \textbf{Relative speed} \\
    \midrule
    BM25             & \checkmark & \texttt{rank-bm25}             & No  & Fastest   \\
    TF-IDF           & \checkmark & \texttt{scikit-learn}          & No  & Fast      \\
    Embedding        & $\sim$     & \texttt{sentence-transformers} & Yes & Moderate  \\
    Cross-Encoder    & $\sim$     & \texttt{sentence-transformers} & Yes & Slow      \\
    \bottomrule
  \end{tabular}}
  \caption{Ranker method comparison. ``Deterministic'' marks methods whose
           output is identical across runs given the same input.
           Embedding and cross-encoder results may vary across model versions.}
  \label{tab:ranker_comparison}
\end{table}

\section{Latency Benchmark}
\label{app:latency}

\begin{table}[h]
  \centering
  \small
  \resizebox{\columnwidth}{!}{%
  \begin{tabular}{lccc}
    \toprule
    \textbf{Chunking (6k budget)}  & \textbf{1 chunk} & \textbf{6 chunks} & \textbf{11 chunks} \\
    \midrule
    gpt-4o-mini   & 0.7 ms & 6.6 ms & 14.2 ms \\
    llama-4-scout & 0.7 ms & 7.6 ms & 13.9 ms \\
    \midrule
    \textbf{Selection}  & \textbf{5 cand.} & \textbf{10 cand.} & \textbf{20 cand.} \\
    \midrule
    First / Reader-conf.  & $<$0.01 ms & $<$0.01 ms & $<$0.01 ms \\
    BM25          & 0.06 ms & 0.08 ms & 0.12 ms \\
    TF-IDF        & 1.8 ms & 1.8 ms & 2.2 ms \\
    Embedding     & 28.4 ms & 30.4 ms & 32.2 ms \\
    Cross-Encoder & 28.5 ms & 30.5 ms & 31.7 ms \\
    \bottomrule
  \end{tabular}}
  \caption{Wall-clock latency (CPU only, median of 100 runs, chunkrank 2.0.0).
           Chunking scales with the number of chunks and is essentially
           model-independent (tiktoken vs.\ Hugging Face tokenizer overhead is
           small); selection scales with candidate count. The default \texttt{first}
           and \texttt{confidence} selectors are effectively free; only the neural
           methods add tens of milliseconds. Reader/LLM inference itself (not shown)
           dominates total pipeline latency by orders of magnitude.}
  \label{tab:latency}
\end{table}

\section{Ranking Without Positional Bias (HotpotQA)}
\label{app:hotpot}

Full results for the positional check (Section~\ref{sec:experiments}). We use
HotpotQA \cite{yang2018hotpotqa} in the distractor setting as a
\emph{single-document} diagnostic: we concatenate the ten distractor paragraphs
into one document (so the answer-bearing paragraph sits at mean index 4.6, only
$10\%$ at position 0) and ask whether a strong first-chunk baseline is merely
positional. We keep the 400 examples whose answer is a non-yes/no span appearing
verbatim in the context, split at a 256-token budget ($\sim$8 chunks each), and run
the same harness as Table~\ref{tab:results} over the non-empty candidate set
(random-baseline seed 42). Even here, where the answer is not front-loaded, no
content-based method significantly beats first-non-empty (Table~\ref{tab:hotpot}):
embedding, cross-encoder, and reader-confidence edge it by $0.4$--$1.0$ EM, but
none of the differences is significant (paired bootstrap $p>0.4$). This rules out
the positional explanation: abstention, not answer position, is what makes
first-non-empty strong. We do not evaluate multi-hop reasoning, which post-chunk
selection cannot perform and which caps the absolute scores (Limitation~1).

\begin{table}[h]
  \centering
  \small
  \begin{tabular}{lcc}
    \toprule
    \textbf{Method} & \textbf{EM} & \textbf{F1} \\
    \midrule
    First (non-empty) & 21.8 & 27.9 \\
    Last              & 20.0 & 26.6 \\
    Random            & 20.5 & 26.9 \\
    \midrule
    BM25              & 21.8 & 27.8 \\
    TF-IDF            & 21.8 & 28.0 \\
    Embedding         & 22.2 & 29.0 \\
    Cross-Encoder     & 22.5 & \textbf{29.1} \\
    Reader-conf.      & \textbf{22.8} & 28.7 \\
    \midrule
    Oracle            & 27.8 & 35.0 \\
    \bottomrule
  \end{tabular}
  \caption{HotpotQA (distractor) EM/F1, 400 examples, 256-token budget, over the
           non-empty candidate set. \textbf{Bold} marks the best non-oracle result
           per column, but no method significantly beats first-non-empty (all
           paired-bootstrap $p>0.4$). Even with answers that are not front-loaded,
           first-non-empty is not beaten, so the baseline's strength is due to
           reader abstention, not answer position.}
  \label{tab:hotpot}
\end{table}

\end{document}